\documentclass[runningheads]{llncs}
\usepackage{graphicx}
\usepackage[symbol*]{footmisc}
\usepackage{subfigure}
\usepackage{amsmath,amssymb}

\begin{document}
%===========================================================

\title{Believe It or Not, We Know What You Are Looking at!} % Replace your paper's title here
\titlerunning{Believe It or Not, We Know What You Are Looking at!} % Replace an abstracted version of your paper's title here

%===========================================================

\author{Dongze Lian\thanks{The authors contribute equally.}\orcidID{0000-0002-4947-0316} \and
Zehao Yu$^*$\orcidID{0000-0002-6559-9830} \and
Shenghua Gao\thanks{Corresponding author.}\orcidID{0000-0003-1626-2040}}

\authorrunning{Lian et al.}

\institute{School of Information Science and Technology, ShanghaiTech University\\
\email{\{liandz, yuzh, gaoshh\}@shanghaitech.edu.cn}}

\maketitle

\begin{abstract}
	By borrowing the wisdom of human in gaze following, we propose a two-stage solution for gaze point prediction of the target persons in a scene. Specifically, in the first stage, both head image and its position are fed into a gaze direction pathway to predict the gaze direction, and then multi-scale gaze direction fields are generated to characterize the distribution of gaze points without considering the scene contents. In the second stage, the multi-scale gaze direction fields are concatenated with the image contents and fed into a heatmap pathway for heatmap regression. There are two merits for our two-stage solution based gaze following: i) our solution mimics the behavior of human in gaze following, therefore it is more psychological plausible; ii) besides using heatmap to supervise the output of our network, we can also leverage gaze direction to facilitate the training of gaze direction pathway, therefore our network can be more robustly trained. Considering that existing gaze following dataset is annotated by the third-view persons, we build a video gaze following dataset, where the ground truth is annotated by the observers in the videos. Therefore it is more reliable. The evaluation with such a dataset reflects the capacity of different methods in real scenarios better. Extensive experiments on both datasets show that our method significantly outperforms existing methods, which validates the effectiveness of our solution for gaze following. Our dataset and codes are released in \url{https://github.com/svip-lab/GazeFollowing}.
	
\end{abstract}

%===========================================================
\section{Introduction}
\label{sec:introduction}
\begin{figure}
	\centering
	\subfigure[]{
		\begin{minipage}[b]{0.22\textwidth}
			\includegraphics[width=0.9\linewidth]{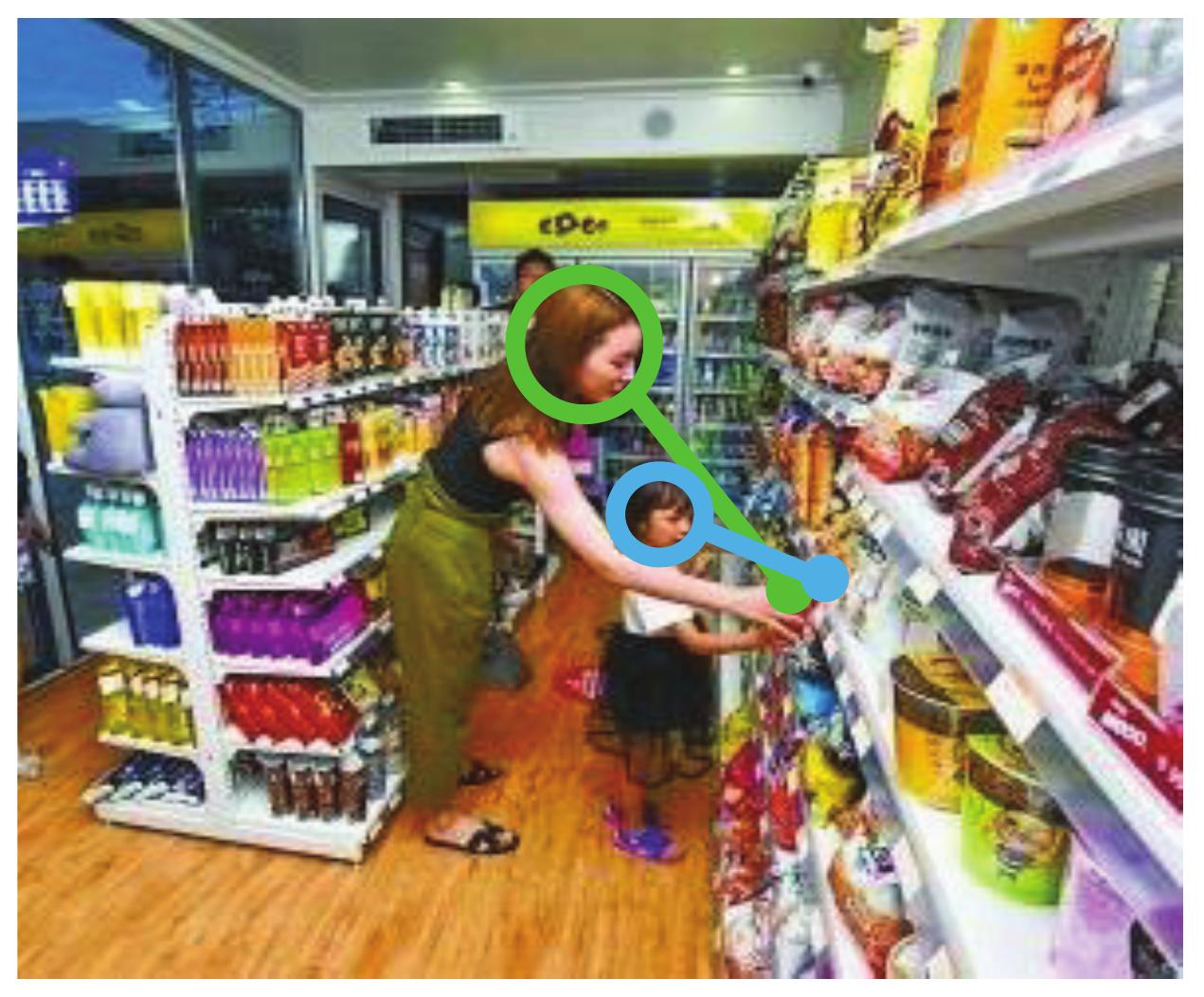}
		\end{minipage}
	}
	\subfigure[]{
		\begin{minipage}[b]{0.22\textwidth}
			\includegraphics[width=0.9\linewidth]{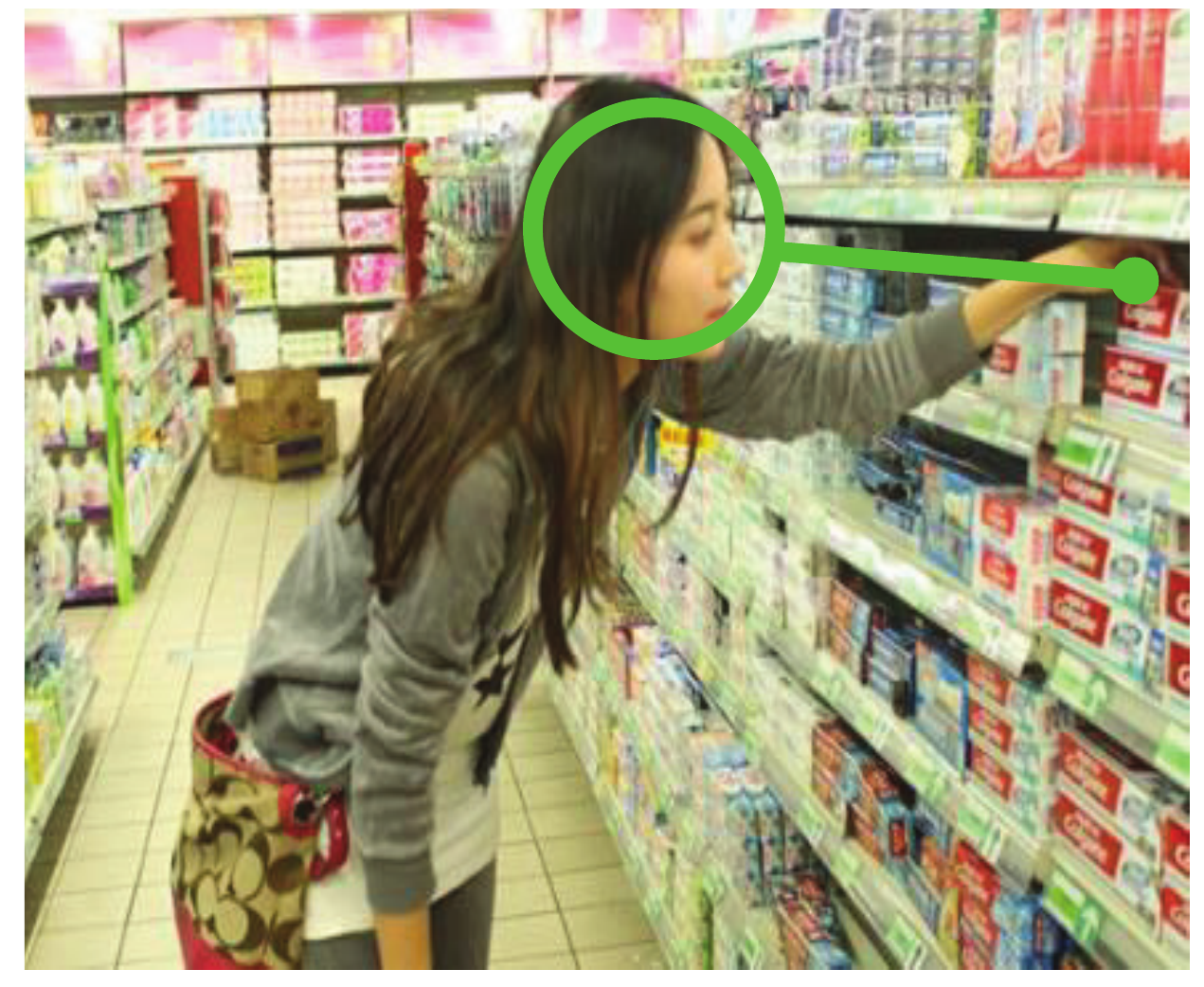}
		\end{minipage}
	}
	\subfigure[]{
		\begin{minipage}[b]{0.22\textwidth}
			\includegraphics[width=0.98\linewidth]{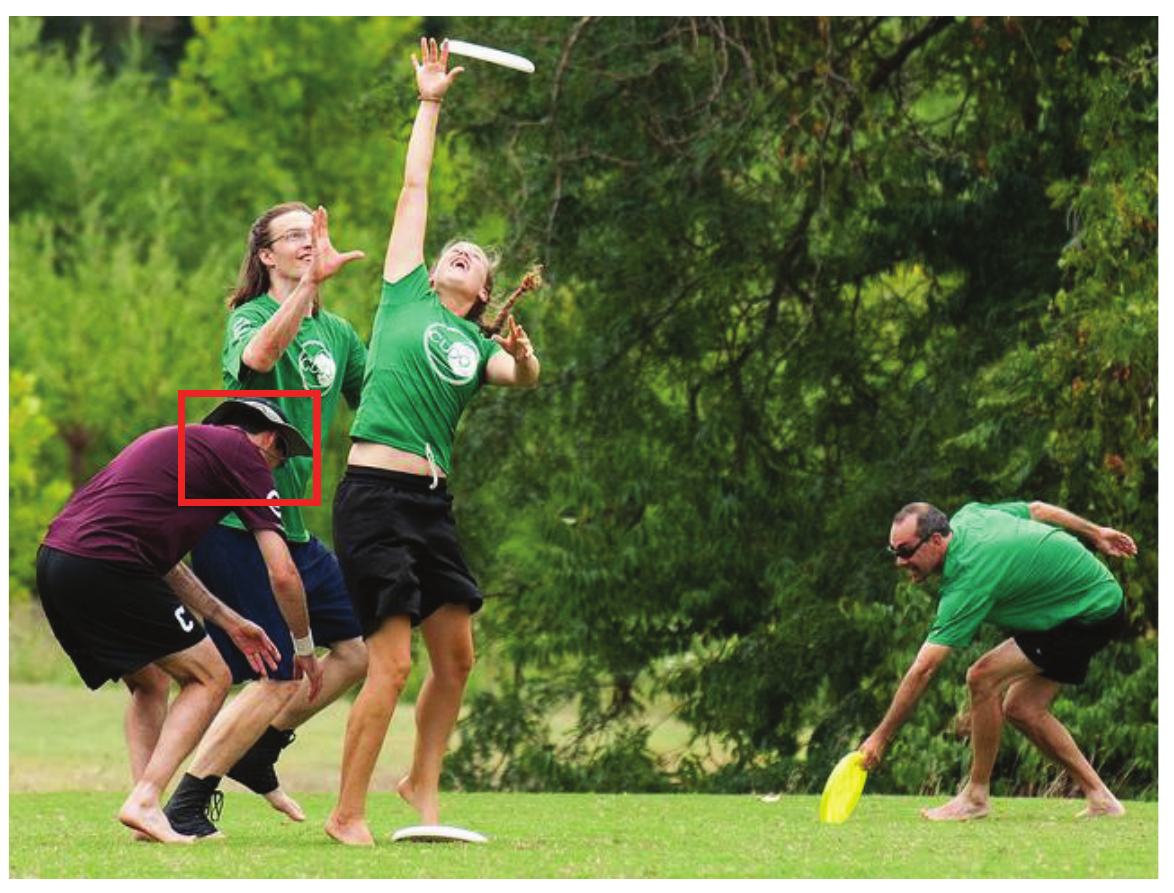}
		\end{minipage}
	}
	\subfigure[]{
		\begin{minipage}[b]{0.22\textwidth}
			\includegraphics[width=1\linewidth]{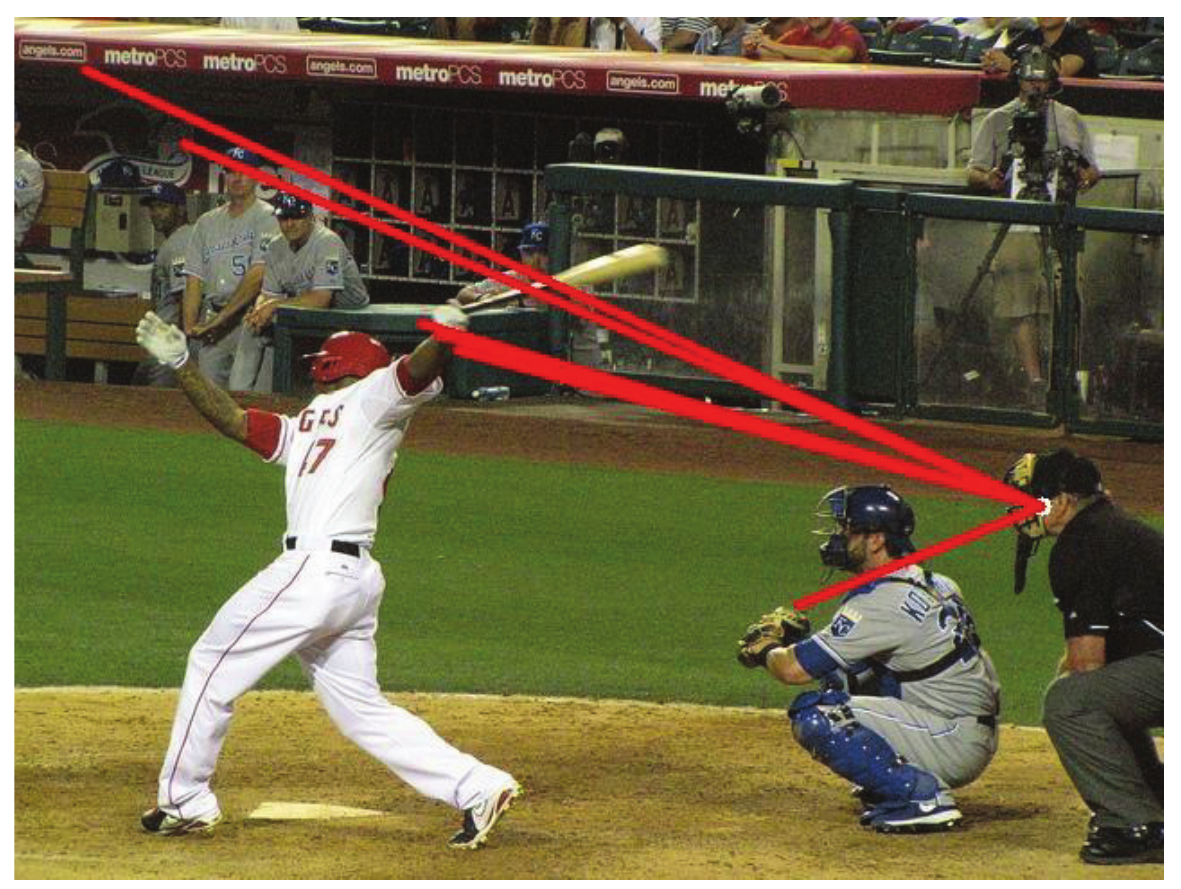}
		\end{minipage}
	}
	\caption{(a) and (b) show the application of gaze following in supermarket scenario. (c) and (d) show the challenges in gaze following (The self-occluded head and ambiguity of gaze point).}
	\label{fig:introduction}
\end{figure}

Gaze following is a task of following other people's gaze in a scene and inferring where they are looking \cite{nips15_recasens}. It is important for understanding the behavior of human in human-human interaction and human-object interaction. For example, we can infer the intention of persons based on their gaze points in human-human interaction. In new retailing scenario, we can infer the interest of the consumers in different products based on their eyes contact with those products (as shown in Figure \ref{fig:introduction} (a) (b)), and infer what kind of information (ingredients of the food, the price, expire data, \emph{etc.}) attracts the consumers' attention most. Although gaze following is of vital importance, it is extremely challenging because of the reasons below: firstly, actually inferring the gaze point requires the depth information of the scene, head pose and eyeball movement \cite{Zhu_2017_ICCV,xiong2014eye}, nevertheless it is hard to infer the depth of scene with a monocular image. Further, head pose and eyeball movements are not easy to be estimated because of occlusion (usually self-occlusion), as shown in Figure \ref{fig:introduction} (c); secondly, ambiguity exists for gaze point estimated by different third-view observers with a single view image, as shown in Figure \ref{fig:introduction} (d); thirdly, the gaze following involves the geometric relationship understanding between target person and other objects/persons in the scene as well as scene contents understanding, which is a difficult task.

To tackle these problems, early works usually simplify the setting to avoid the handicaps for general gaze following, for example, making assumption that face is available for better head pose estimation \cite{zhu2012face}, with multiple inputs for depth inference \cite{mukherjee2015deep}, with eye tracker for ground truth annotation or restricting the application scenario to people looking at each other for disambiguation. However, these simplifications restrict the applications of general gaze following. Recently, Recasens \emph{et al.} propose to study general gaze following under most of general settings \cite{nips15_recasens}. Specifically, they propose a two-pathway method (a gaze pathway and a saliency pathway) based deep neural networks for gaze following. However, in their solution, the saliency pathway and gaze pathway are independent of each other. For a third-view person, when he/she infers the gaze point of a target person, he/she infers the gaze direction first based on head pose, and then estimates the gaze point from the scene contents along the gaze direction, where gaze point denotes the position that one person is looking at in the image, and gaze direction means the direction from head position to gaze point in this paper. In other words, the saliency pathway relies on gaze direction estimation, which is neglected in \cite{nips15_recasens}.  

In this paper, we propose a two-stage solution to mimic the behavior of a third-view person for gaze following. Specifically, in the first stage, we leverage a gaze direction pathway to predict the gaze direction based on the head image and head position of target person. There are two motivations for such gaze direction pathway: firstly, it is more natural to infer the gaze direction rather than gaze point merely based on head image and head position; secondly, since the gaze direction can be inferred in the training phase, thus we can introduce a loss w.r.t. gaze direction to facilitate the learning of this gaze direction pathway. Next, we encode the predicted gaze direction as the multi-scale gaze direction fields. In the second stage, based on the gaze direction and the context information of the objects along the gaze direction, we can estimate a heatmap through a heatmap pathway. In this stage, we concatenate the multi-scale gaze direction fields with the original image as the input of the heatmap pathway for heatmap estimation.

A proper dataset is important for the evaluation of gaze following. The only existing gaze following dataset (GazeFollow dataset \cite{nips15_recasens}) is annotated by the third-view persons. In this paper, to evaluate the performance for real problem, we build a video-based gaze following dataset, named Daily Life Gaze dataset (DL Gaze). Particularly, we have 16 volunteers to freely move in 4 different indoor scenes, including working office, laboratory, library, corridor in the building. During the period, they can talk, read books, use their mobile phones, or freely look at other places in the scene. We record the video for them and ask the volunteer to annotate where they look later. There are 95,000 frames in total. Compared with GazeFollow, the ground truth annotated by the persons in the video is more reliable than that annotated by third-view workers. Further, it is a video-based gaze following dataset and records the gaze following for real scenes. Therefore the evaluation of gaze following on this dataset reflects the performance of different methods for real problem. 

The main contributions of our paper are summarized as follows: i) we propose a two-stage solution for gaze following task. Our network architecture is inspired by the behavior of human in gaze following, therefore it is more psychological plausible; ii) we use ground truth to supervise the learning of both stages, consequently facilitates the network training. In addition, we introduce multi-scale gaze direction fields for attention prediction, which further improves the performance of gaze following; iii) we collect a video-based gaze following dataset (DL Gaze), with the ground truth annotated by the persons in the video. Therefore the evaluation on this dataset reflects the real performance for gaze following in real problem; iv) Extensive experiments on both datasets validate the effectiveness of our solution for gaze following.

\section{Related Work}
\label{sec:related_work}

\textbf{Gaze following.}
Previous work about gaze following paid attention to restricted scenes, which added some priors for specific applications. In \cite{zhu2012face}, a face detector was employed to extract face, which was limited for the people looking away from the camera. \cite{marin2014detecting} detected whether people were looking at each other in a movie, which was helpful for interaction. Eye tracker was utilized to predict the next object in order to improve action recognition in \cite{fathi2012learning}. \cite{parks2015augmented} only estimated the gaze direction from head position, but not the specific gaze point. These methods were applied to a particular scene. Recent works \cite{nips15_recasens,mukherjee2015deep,recasens2017following} focused on general gaze following, which had wider applications. Given a single picture containing one or more people, the gaze points of some people in the image were estimated, without any restrictions in \cite{nips15_recasens}. Some extensive works \cite{mukherjee2015deep,recasens2017following} focused on multi-modality image or predicted gaze point in videos. The RGB-D image was introduced to predict gaze in images and videos \cite{mukherjee2015deep} because the multi-modality data provided 3D head pose information in order to find more accurate gaze point. In \cite{recasens2017following}, the cross-frame gaze point in videos could be predicted for the people in a frame.

\textbf{Eye tracking.}
Eye tracking is strongly related to gaze following. Different from gaze following, eye tracking technology inferred which direction or which point on the screen one person was looking at \cite{zhang2015appearance}. Previous work \cite{zhu2005eye,hennessey2006single} built the geometry model to infer the gaze point on the screen target. Recently, many appearance-based methods \cite{krafka2016eye,Zhu_2017_ICCV} solved the problem by learning a complex function from the eye images to gaze point, which needed large-scale dataset. These methods took the eye images and face image as inputs because gaze direction could be determined according to the eye movement and head pose \cite{Zhu_2017_ICCV}. However, the eye images could not be utilized to predict gaze point because they were occluded or very noisy in gaze following. Thus, gaze following direction is almost obtained from the head image.

\textbf{Saliency.}
Saliency detection and gaze following are two different tasks \cite{nips15_recasens,recasens2017following} even though they were closely related. Saliency detection predicts fixation map from observers out of the original images \cite{itti2001computational,judd2009learning,leifman2017learning}. Gaze following in image predicts the position that people in a scene were looking at. Previous works about saliency prediction considered the low-level features and saliency maps at different scales \cite{itti1998model}. Subsequently, the features from different levels were combined to model a bottom-up, top-down architecture \cite{judd2009learning}. Recently, deep neural networks have been applied to saliency prediction and achieve great success \cite{kummerer2014deep,pan2017salgan}. However, the object in the gaze point region may be not salient, which reveals that it is hard to find the gaze point through a saliency algorithm directly.

\section{Approach}
\label{sec:approach}

\begin{figure*}[h]
	\centering
	\includegraphics[width=1\linewidth]{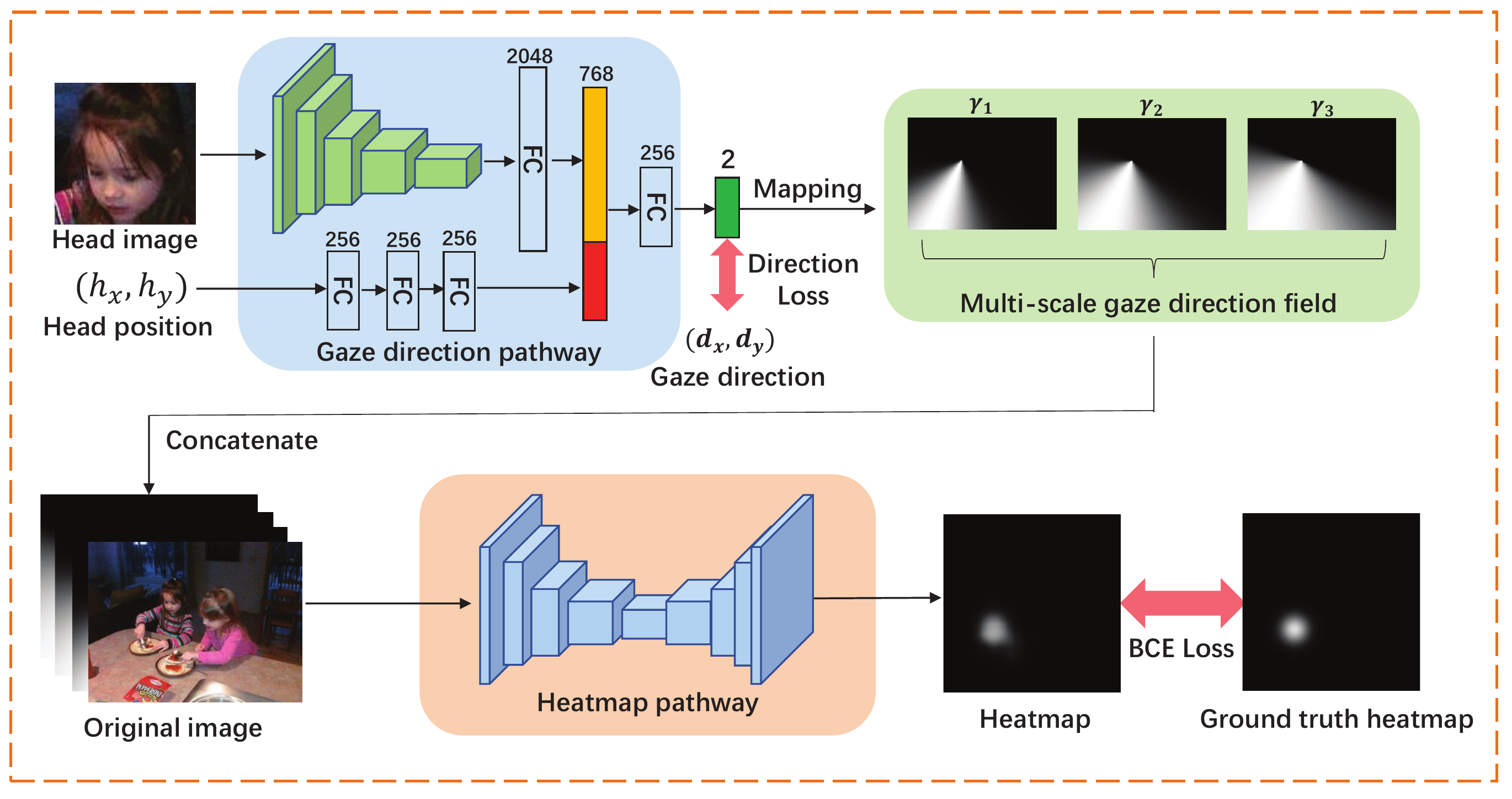}
	\caption{The network architecture for gaze following. There are two modules in this network: gaze direction pathway and heatmap pathway. In the first stage, a coarse gaze direction is predicted through gaze direction pathway, and then it is encoded as multi-scale gaze direction fields. We concatenate the multi-scale fields and the original image to regress heatmap of final gaze point through heatmap pathway.}
	\label{fig:architecture}
\end{figure*}

Inspired by the behavior of human in gaze following, we propose a two-stage solution. Specifically, when a third-view person estimates the gaze of the target person, he/she first estimates the gaze direction of the target based on the head image, then the gaze point is predicted based on the scene content along the gaze direction. Similarly, we feed the head image and its position in the image into a gaze direction pathway for gaze direction prediction in the first stage, and then the multi-scale gaze direction fields are encoded. In the second stage, the gaze direction fields are concatenated with the original image as the input of heatmap pathway for heatmap regression. It is worth noting that all components in our network are differentiable and the whole network can be trained with an end-to-end learning. The  network architecture is shown in Figure \ref{fig:architecture}.

\subsection{Gaze direction pathway}
Gaze direction pathway takes head image and head position as inputs for gaze direction prediction. We feed the head image into a ResNet-50 for feature extraction, and then concatenate head features with head position features encoded by a network with three fully connected layers for gaze direction prediction. Different from work of Recasens \emph{et al.} \cite{nips15_recasens}, which takes head image and head position for gaze mask prediction, our network only estimates the gaze direction. There are two reasons accounting for our solution. Firstly, it is easier to infer the gaze direction than gaze mask merely based on head image and its position. Secondly, we can use gaze direction to supervise the learning of gaze direction pathway to make it more robustly trained. It is also worth noting that the predicted gaze direction would be used to generate gaze direction fields, which is further used for heatmap regression in the heatmap pathway, and the optimization of heatmap would also update the parameters in the gaze direction pathway.

\subsection{Gaze direction field}

\begin{figure}
	\centering
	\subfigure[]{
		\begin{minipage}[b]{0.45\textwidth}
			\includegraphics[width=1\linewidth]{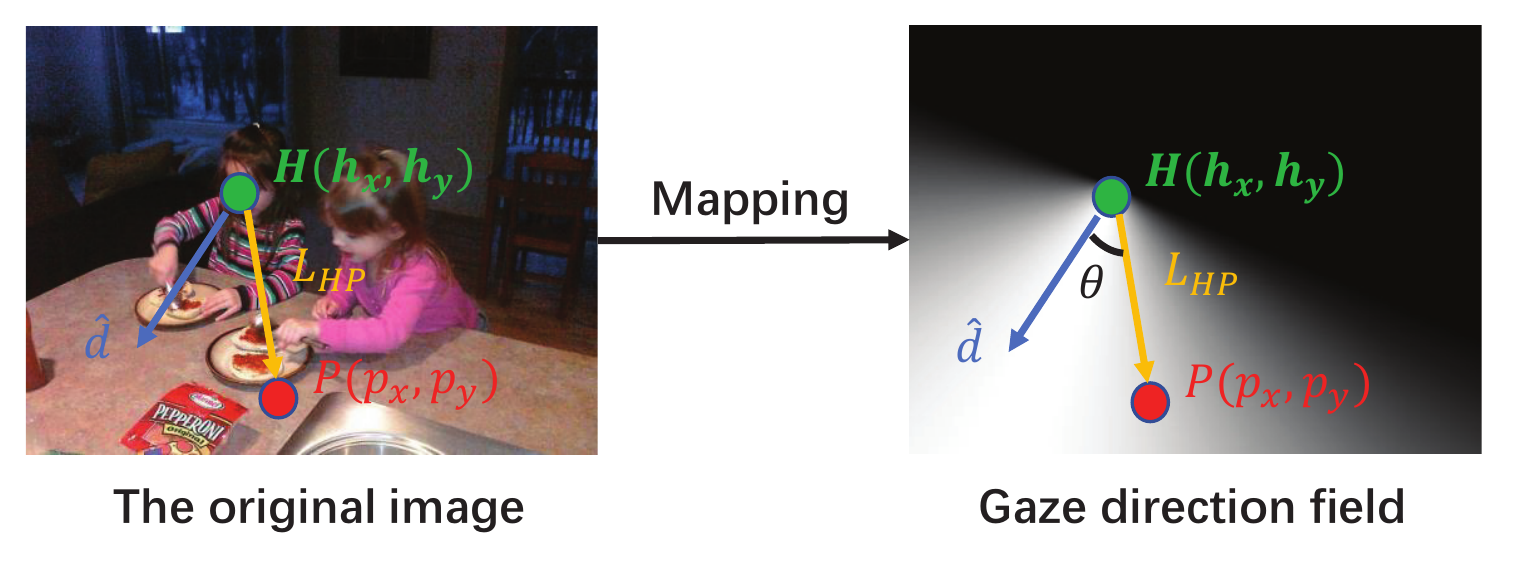}
		\end{minipage}
	}
	\subfigure[]{
		\begin{minipage}[b]{0.45\textwidth}
			\includegraphics[width=1\linewidth]{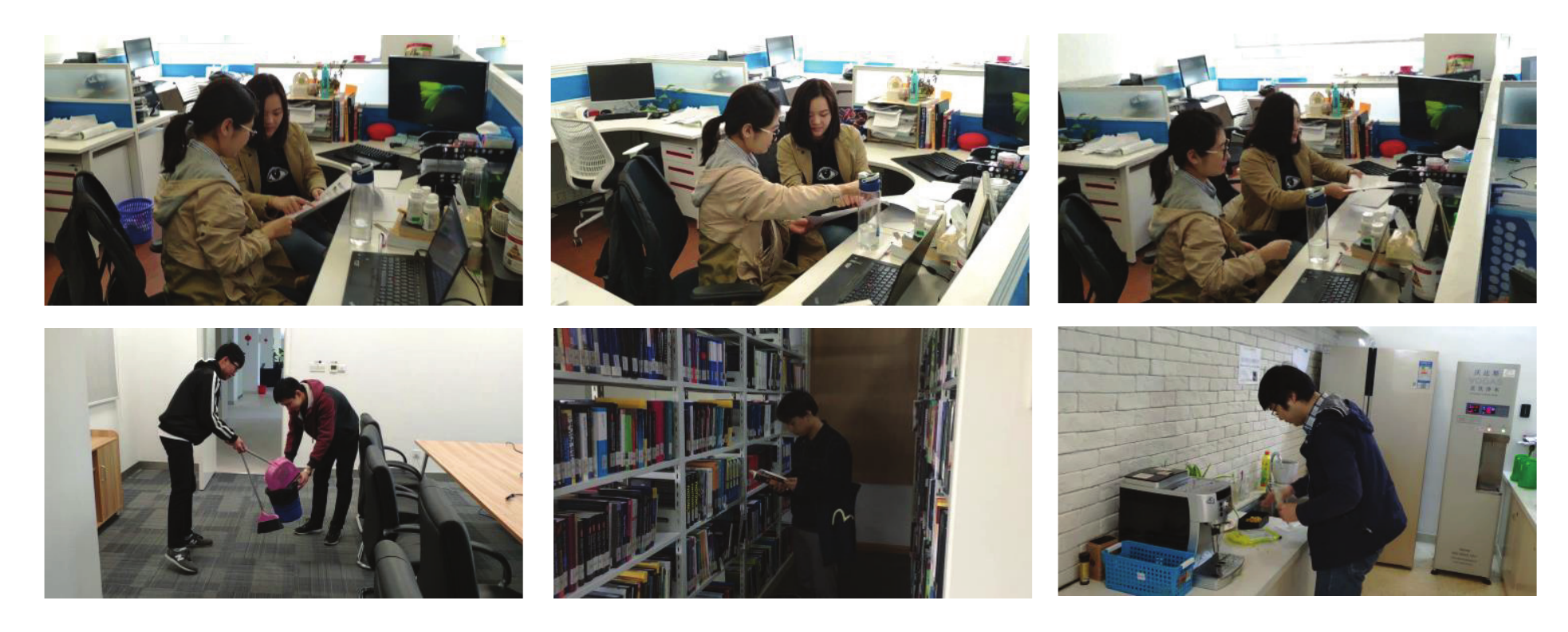}
		\end{minipage}
	}
	\caption{(a) The original image: the blue line shows gaze direction of the left girl inside the image, and the green dot shows the head position. Gaze direction field, which measures the probability of each point being gaze point with cosine function between the line direction of $L_{HP}$ and predicted gaze direction $\hat{d}$. (b) Our DL Gaze dataset.}
	\label{fig:gaze_field}
\end{figure}

Once the gaze direction is estimated, the gaze point is likely to be along the gaze direction. Usually, the field of view (FOV) of the target person is simplified as a cone with the head position as the apex of the cone. So given a point $P=(p_x,p_y)$, if we do not consider the scene contents, then the probability of the point $P$ being the gaze point should be proportional to the angle $\theta$ between line $L_{HP}$ and predicted gaze direction, here $H=(h_x, h_y)$ is the head position, as shown in Figure \ref{fig:gaze_field} (a). If $\theta$ is small, then the probability of the point being gaze point is high, otherwise, the probability is low. We utilize the cosine function to describe the mapping from the angle to the probability value. We denote the probability distribution of the points being gaze point without considering the scene contents as the \textbf{gaze direction field}. Thus, gaze direction field is a probability map, where intensity value of each point shows the probability that this point is the gaze point. Its size is the same as the scene image.

Particularly, the line direction of $L_{HP}$ can be calculated as follows:
\begin{equation}
G = (p_x-h_x, p_y-h_y)
\end{equation}
Given an image with size $W\times H$ (here $W$ and $H$ are the width and height of image, respectively), we denote the predicted gaze direction as $ \hat{d} = (\hat{d_x}, \hat{d_y})$, then the probability of the point $P$ being the gaze point can be calculated as follows:
\begin{equation}
Sim(P) = \max\Big(\dfrac{\langle G, \hat{d} \rangle}{|G||\hat{d}|}, 0\Big)
\end{equation}
Here we let the probability of $P$ being gaze point to be 0 when the angle between gaze direction and line $L_{HP}$ is larger than $90^{\circ}$, which means the real gaze direction should not contradict with the predicted gaze direction. We depict the calculation of gaze direction field.

If the predicted gaze direction is accurate, it is desirable that the probability distribution is sharp along the gaze direction, otherwise, it is desirable that the probability changes smoothly. In practice, we leverage multi-scale gaze direction fields with different sharpness for heatmap prediction. Specifically, we use the following way to control the sharpness of the gaze direction field:
\begin{equation}
Sim(P,\gamma) = [Sim(P)]^{\gamma}
\end{equation}
Here $\gamma$ controls aperture of the FOV cone. Larger $\gamma$ corresponds to a FOV cone with smaller aperture, as shown in Figure \ref{fig:architecture}. In our implementation, considering the change rate of $Sim(P,\gamma)$, we empirically set $\gamma_1 = 5, \gamma_2 = 2, \gamma_3 = 1$. More details about $\gamma$ can be found in the supplementary material.

It also worth noting that the gaze direction fields are differentiable w.r.t. the network parameters of gaze direction pathway, so that the whole architecture can be trained with an end-to-end learning strategy.

\subsection{Heatmap pathway}

Gaze direction fields encode the distribution of gaze points inferred from gaze direction, together with scene contents, we can infer the gaze point. Specifically, we concatenate the original image and the multi-scale gaze direction fields, and feed them into a heatmap pathway for heatmap regression. The point corresponding to the maximum value of the heatmap is considered as the final gaze point. In practice, we leverage a feature pyramid network (FPN) \cite{lin2017feature} for the heatmap pathway in light of its success in object detection. The last layer of heatmap pathway is followed with a Sigmoid activation function, which guarantees the probability of each pixel falls into $[0, 1]$.

There are two reasons to predict probability heatmap instead of a direct gaze point coordinate:
\begin{itemize}
	\item As pointed in \cite{tompson2014joint}, mapping from image to the coordinates of gaze point directly is a highly non-linear function. Compared with gaze point estimation, heatmap prediction is more robust, which means even some entries of heatmap are not accurately predicted, the gaze point prediction based on heatmap can still be correct. Thus heatmap regression is more commonly used in many applications, including pose estimation \cite{pfister2015flowing} and face alignment. The experimental results in section \ref{sec:experiments} also validate the advantage of heatmap regression over gaze point regression.
	\item Gaze following in an image is sometimes ambiguous \cite{nips15_recasens} due to the lack of the ground truth. Different workers may vote for different gaze points, especially when the eye images are invisible, the head image is low-resolution or occluded. Thus, the gaze point is usually multimodal, and the output of network is expected to support the multimodal prediction. Heatmap regression satisfies such requirement. 
\end{itemize}

Following \cite{pfister2015flowing}, the heatmap of ground truth gaze point is generated by centering a Gaussian kernel at the position of gaze point as follows:
\begin{equation}
H(i, j) = \dfrac{1}{\sqrt{2\pi} \sigma }e^{-\dfrac{(i - g_x)^2 + (j - g_y)^2}{2\sigma ^2}}
\end{equation}
where $g = (g_x, g_y)$ and $H(i, j)$ are the ground truth gaze point and its heatmap, respectively. $\sigma$ is the variance of Gaussian kernel. We empirically set $\sigma = 3$ in our implementation.

\subsection {Network training}

The inputs of our network consist of three parts: head image, head position and the original image. The head and original image are resized to $224\times 224$, and the head position is the coordinate when the original image size is normalized to 1 $\times$ 1. The outputs of network consist of two parts: gaze direction and visual attention. The gaze direction is the normalized vector from the head position to gaze point and the visual attention is a heatmap with size $56 \times 56$, whose values indicate the probability that the gaze point falls here.

Specifically, the gaze direction loss is:
\begin{equation}
\ell_d  = 1-\dfrac{\langle d, \hat{d} \rangle}{|d||\hat{d}|}
\end{equation}
where $d$ and $\hat{d}$ are the ground truth and predicted gaze direction, respectively.

We employ the binary cross entropy loss (BCE Loss) for heatmap regression, which is written as follows:
\begin{equation}
\ell_h = -\dfrac{1}{N}\sum_{i=1}^N H_i \log(\hat{H_i}) + (1-H_i)\log(1-\hat{H_i})
\end{equation}
where $H_i$ and $\hat{H_i}$ are the $i$-th entry of ground truth heatmap and predicted visual heatmap, respectively. $N$ is with the size $56 \times 56$.

The whole loss function consists of gaze direction loss and heatmap loss:
\begin{equation}
\ell = \ell_d + \lambda \ell_h
\end{equation}

where $\lambda$ is the weight to balance $\ell_d$ and $\ell_h$. We set $\lambda$ = 0.5 in our experiments.

\section{Experiments}
\label{sec:experiments}
\subsection{Dataset and evaluation metric}
\subsubsection{Dataset.}
The GazeFollow dataset \cite{nips15_recasens} is employed to evaluate our proposed method. The images of this dataset are from different source datasets, including SUN \cite{xiao2010sun}, MS COCO \cite{lin2014microsoft}, Actions 40 \cite{yao2011human}, PASCAL \cite{everingham2010pascal}, ImageNet \cite{russakovsky2015imagenet} and Places \cite{zhou2014learning}, which is challenging due to variety of scenarios and amounts of people. The whole dataset contains 130,339 people and 122,143 images. The gaze points of people are inside the image. There are 4,782 people of dataset used for testing and the rest for training. To keep the evaluation consistency with existing work, we follow the standard training/testing split in \cite{nips15_recasens}.

To validate the performance of different gaze following algorithms for real scenarios, we also build a video-based Daily Life Gaze following dataset (DL Gaze). Specifically, DL Gaze contains the activities of 16 volunteers in 4 scenes, and these scenes include working office, laboratory, library and corridor in the building. They can freely talk, read books, use their mobile phones, and look at other places in the scene, as shown in Figure \ref{fig:gaze_field} (b). We record the video for these volunteers with an iPhone 6s. Then we ask each volunteer to annotate what he/she looks at. Two frames are annotated per second. So the ground truth annotation is more reliable. It worth noting the occlusion also exists and there is severe change of illumination. Therefore our dataset is very challenging. The performance of gaze following in our dataset reflects the capability of gaze following in real scenarios. There are 86 videos, 95,000 frames (30 fps) in total. We test the model trained on GazeFollow with our dataset directly.

\subsubsection{Evaluation metric.}
Following \cite{nips15_recasens}, we employ these metrics (\textbf{AUC}, \textbf{Dist}, \textbf{MDist}, \textbf{Ang}) to evaluate the difference between the predicted gaze points and their corresponding ground truth. Details can be found in the supplementary material. In addition, we also introduce \textbf{Minimum angular error (MAng)}, which measures the minimum angle between the predicted gaze direction and all ground truth annotations:

\subsection{Implementation details}

We implement the proposed method based on the PyTorch framework. In the training stage, We employ a ResNet-50 to extract head image feature and encode the original image feature. The network is initialized with the model pretrained with ImageNet \cite{deng2009imagenet}. When the first stage of training converges, we train the heatmap pathway and finally we finetune the whole network with an end-to-end learning strategy. The hyper-parameters of our network are listed as follows: batch size (128), learning rate ($1e^{-4}$), weight decay (0.0005). Adaptive moment estimation (Adam) algorithm \cite{kingma2014adam} is employed to train the whole network.

\subsection{Performance evaluation}
We compare our proposed method with the following state-of-the-art gaze following methods: 
\begin{itemize}
	\item Judd \emph{et al.} \cite{judd2009learning}: Such a method uses a saliency model as a predictor of gaze and the position with maximum saliency value is used as predicted gaze point inside the image.
	\item SalGAN \cite{pan2017salgan}: SalGAN \cite{pan2017salgan} is the latest saliency method, and it takes the original image as input to generate visual heatmap. The position of maximum in visual attention is regarded as gaze point.
	\item SalGAN for heatmap: We replace the FPN with SalGAN in heatmap pathway, and all the rest components are the same with our method.
	\item Recasens \emph{et al.} \cite{nips15_recasens}: The gaze pathway and saliency pathway are introduced to extract the image and head feature, and both features are fused to get the final gaze point. The supervision is introduced in the last layer. 
	\item Recasens \emph{et al.}*: For a fair comparison, we modify the backbone of Recasens \emph{et al.} \cite{nips15_recasens} from the AlexNet \cite{krizhevsky2012imagenet} to ResNet-50 \cite{he2016deep} to extract head feature and image feature. All other parts remain the same as Recasens \emph{et al.} \cite{nips15_recasens}.
	\item One human \cite{nips15_recasens}: A third-view observer is employed to predict gaze points on the testing set in \cite{nips15_recasens}. It is desirable that machine can achieve the human level performance.
\end{itemize}

\begin{table}[h]
	\centering
	\caption{Performance comparison with existing methods on the GazeFollow dataset. One-scale and multi-scale correspond to the number of gaze direction fields in our model. For one-scale model, $\gamma = 1$.}
	\begin{tabular}{lccccc}
		\hline
		Methods  &~~AUC~~ & ~~Dist~~  & ~~MDist~~ & ~~Ang~~ & ~~MAng~~\\
		\hline
		Center \cite{nips15_recasens} &0.633 &0.313 &0.230 &$49.0^{\circ}$ &-\\
		Random \cite{nips15_recasens} &0.504 &0.484 &0.391 &$69.0^{\circ}$ &-\\
		Fixed bias \cite{nips15_recasens} &0.674 &0.306 &0.219 &$48.0^{\circ}$ &-\\
		SVM + one grid \cite{nips15_recasens} &0.758 &0.276 &0.193 &$43.0^{\circ}$ &-\\
		SVM + shift grid \cite{nips15_recasens} &0.788 &0.268 &0.186 &$40.0^{\circ}$ &-\\
		Judd \emph{et al.} \cite{judd2009learning}&0.711 &0.337 & 0.250 & $54.0^{\circ}$ &-\\
		SalGAN \cite{pan2017salgan} &0.848 & 0.238 & 0.192 & $36.7^{\circ}$  &$22.4^{\circ}$\\
		SalGAN for heatmap  &0.890& 0.181 & 0.107 & $19.6^{\circ}$ & $9.9^{\circ}$\\
		Recasens \emph{et al.} \cite{nips15_recasens} &0.878 & 0.190 & 0.113 & $24.0^{\circ}$ &-\\
		Recasens \emph{et al.}* \cite{nips15_recasens} &0.881 & 0.175 & 0.101 & $22.5^{\circ}$ &$11.6^{\circ}$\\
		One human \cite{nips15_recasens} &0.924 & 0.096 & 0.040 & $11.0^{\circ}$ &-\\
		\hline
		Ours (one-scale) & 0.903 & 0.156 & 0.088 & $18.2^{\circ}$  & $9.2^{\circ}$ \\
		Ours (multi-scale) & \textbf{0.906} & \textbf{0.145} & \textbf{0.081} & $\mathbf{17.6^{\circ}}$ &$\mathbf{8.8^{\circ}}$\\
		\hline
		\hline
	\end{tabular}
	\label{Table:performance comparison}
\end{table}

The experiment results in Table \ref{Table:performance comparison} and Table \ref{Table:our_data} show that our model outperforms all baselines in terms of all evaluation metrics. We also have the following findings: (1) Recasens \emph{et al.}* outperforms Recasens \emph{et al.} shows the importance of the basic network. (2) SalGAN has the better performance than Judd \emph{et al.}, which shows that better saliency detection method agrees with visual attention better. (3) Although employing the same basic network (ResNet-50), our method (one-scale) still achieve the better performance than Recasens \emph{et al.}*, which proves the soundness of our human behavior inspired a two-stage solution for gaze following. (4) The multi-scale model achieves better performance than that of one-scale, which validates the importance of multi-scale fields fusion. (5) Performance on our dataset is worse than that on GazeFollow, which shows the challenge of gaze following in real applications. (6) The improvement of our method over SalGAN for heatmap validates the effectiveness of FPN for heatmap regression.

\begin{figure}[h]
	\centering
	\subfigure{
		\begin{minipage}[b]{0.22\textwidth}
			\includegraphics[width=1\linewidth]{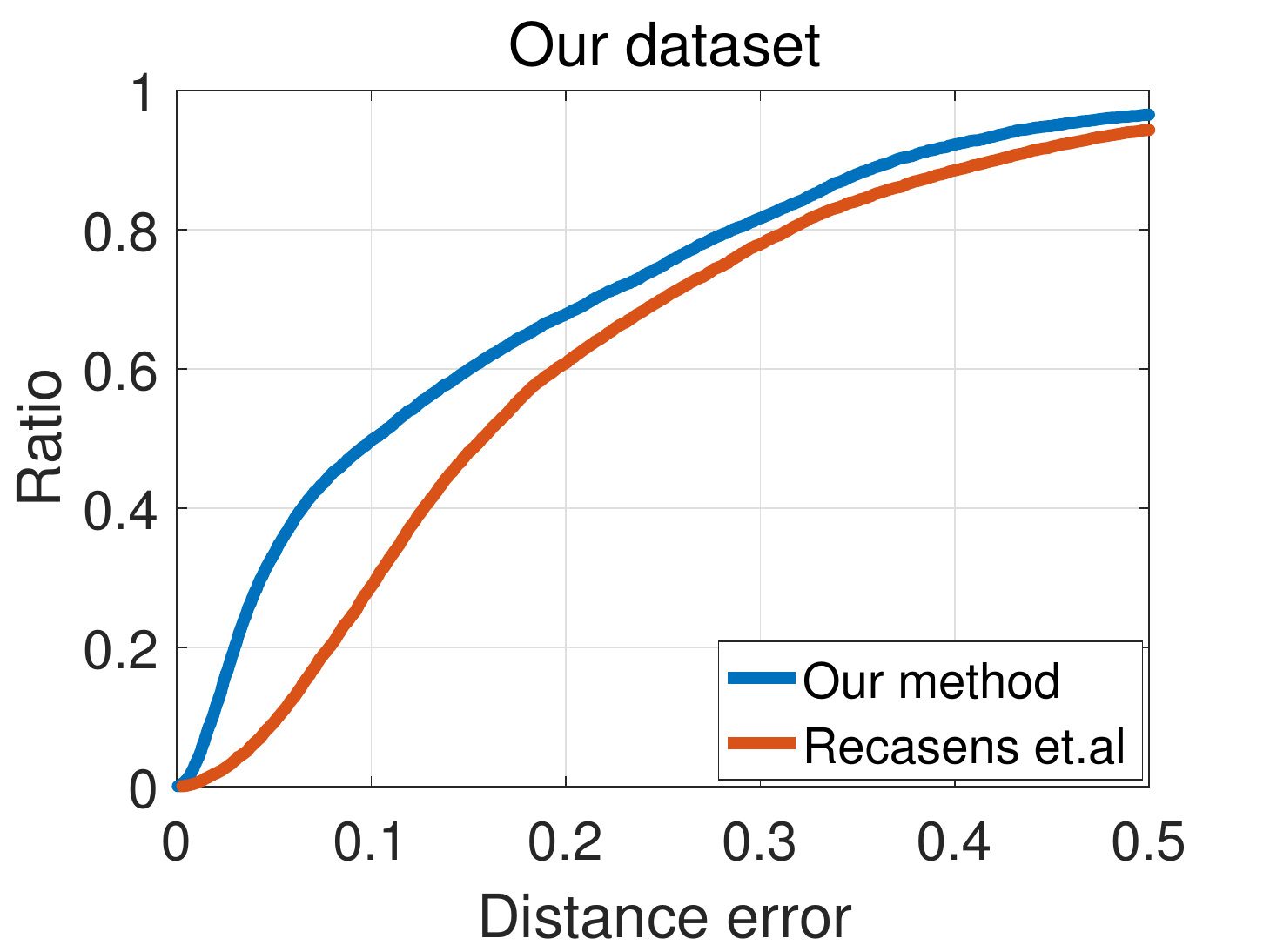}
		\end{minipage}
	}
	\subfigure{
		\begin{minipage}[b]{0.22\textwidth}
			\includegraphics[width=1\linewidth]{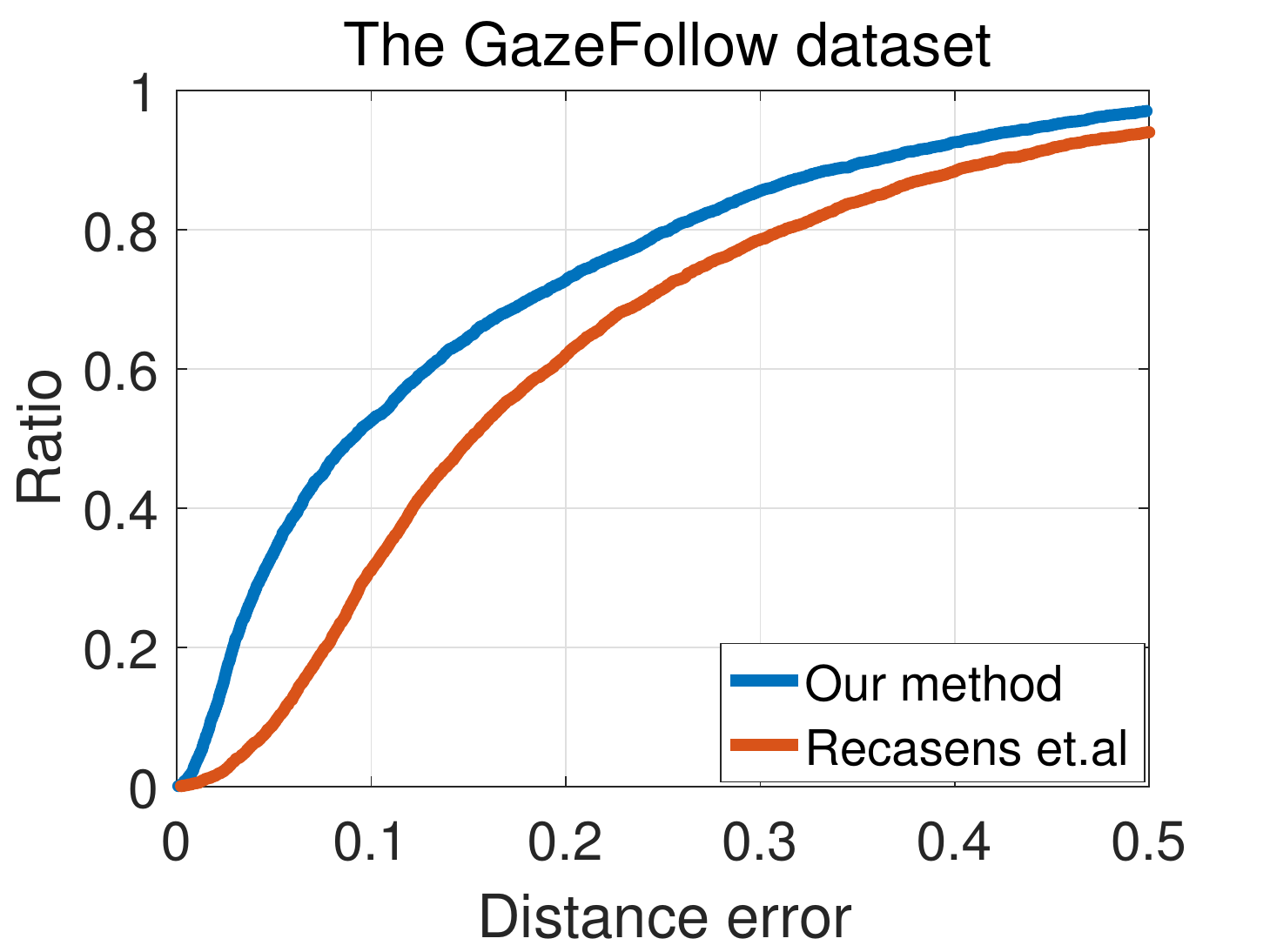}
		\end{minipage}
	}
	\subfigure{
		\begin{minipage}[b]{0.22\textwidth}
			\includegraphics[width=1\linewidth]{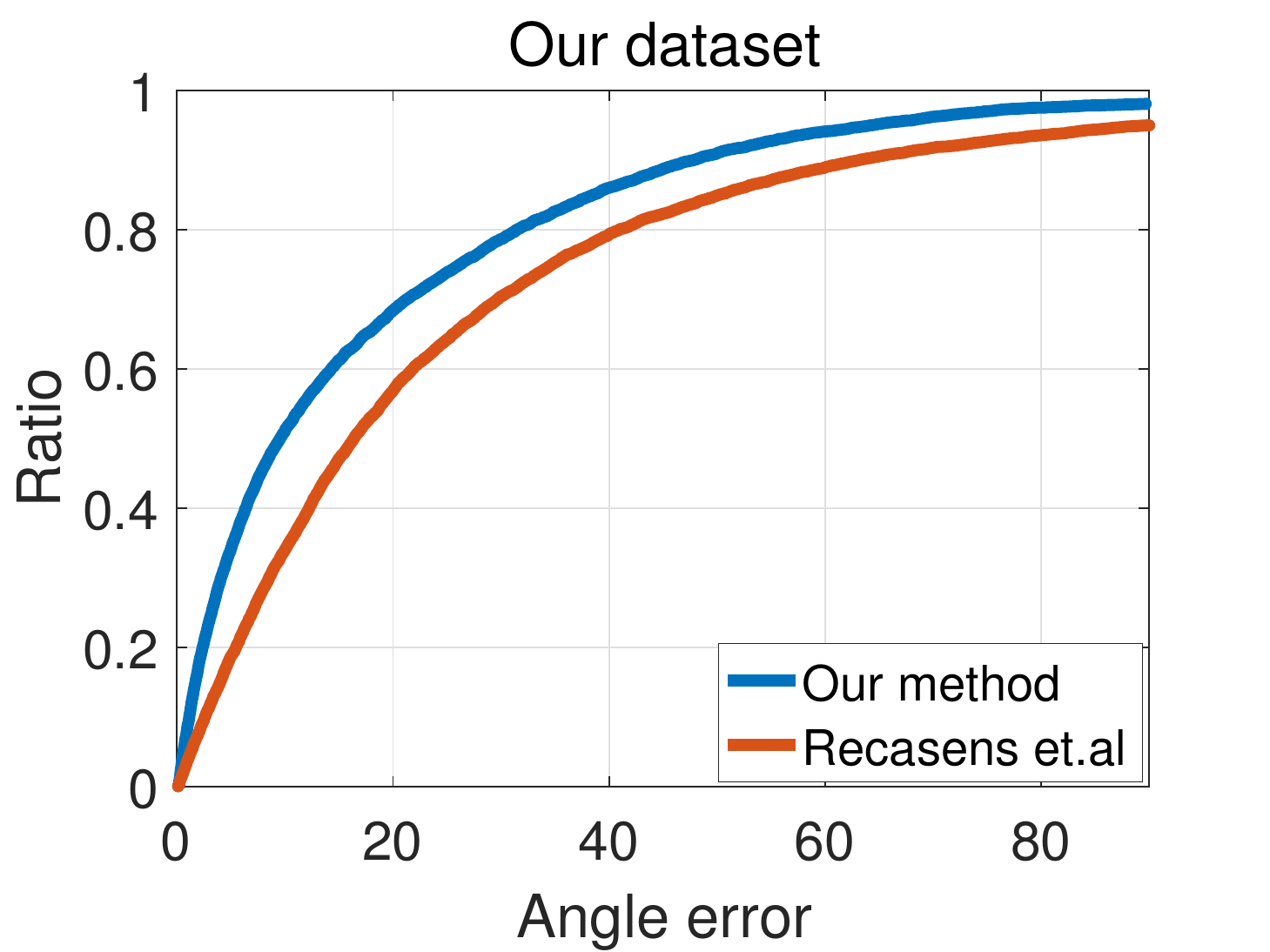}
		\end{minipage}
	}
	\subfigure{
		\begin{minipage}[b]{0.22\textwidth}
			\includegraphics[width=1\linewidth]{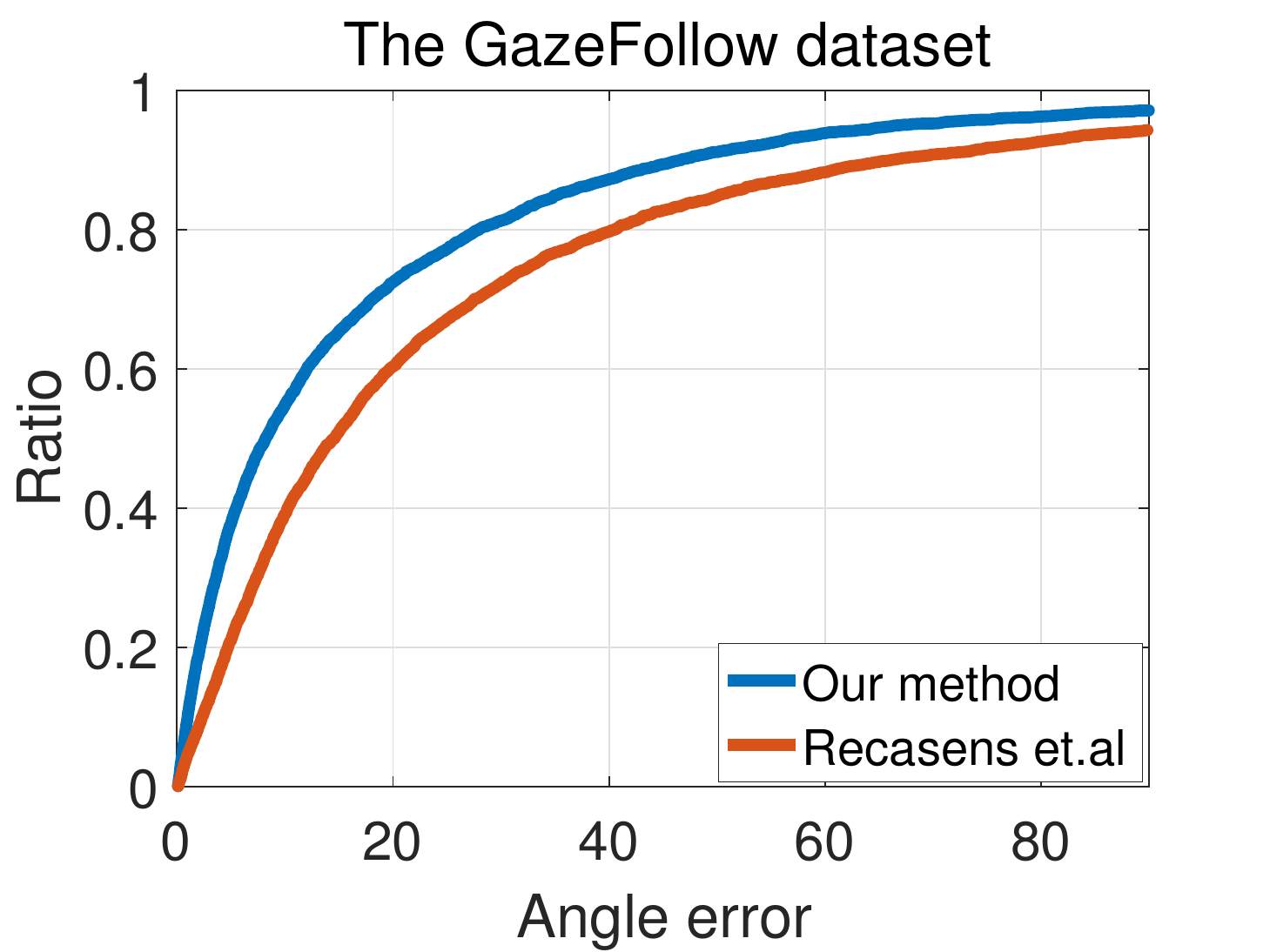}
		\end{minipage}
	}
	\caption{Accumulative error curves of different methods on both datasets.}
	\label{fig:accumulative}
\end{figure}

We further compare our method with Recasens \emph{et al.} using accumulative error curve and the results are shown in Figure \ref{fig:accumulative}. We can see that our method usually achieves better prediction than the work of Recasens \emph{et al.} \cite{nips15_recasens}.

\begin{table}[h]
	\centering
	\caption{Performance comparison with existing methods on our dataset. Each frame only contains one gaze point, so only Dist and Ang are used for performance evaluation.}
	\begin{tabular}{lccccc}
		\hline
		Methods &  ~~Dist~~  & ~~Ang~~ \\
		\hline
		Recasens \emph{et al.} \cite{nips15_recasens}  & 0.203 & $26.9^{\circ}$ \\
		Recasens \emph{et al.}* \cite{nips15_recasens}  & 0.169 & $21.4^{\circ}$ \\
		\hline
		Ours (multi-scale) & 0.157 &$18.7^{\circ}$ \\
		\hline
		\hline
	\end{tabular}
	\label{Table:our_data}
\end{table}

\subsubsection{Ablation study.}
In order to evaluate the effectiveness of every component of different inputs and network. We design the following baselines:
\begin{itemize}
	\item original image:  We directly feed the original image into heatmap pathway for heatmap regression.
	\item original image + ROI head: We directly feed the original image into heatmap pathway for heatmap regression. Further, we directly extract the features corresponding to Region of Interest (ROI, the region of head) from the heatmap pathway and use it for gaze direction regression. Then we train the whole network with multi-task learning.
	\item w/o mid-layer supervision: The gaze direction supervision is removed, and both pathways are trained with an end-to-end learning strategy. Only one-scale gaze direction field is concatenated to the original image.
\end{itemize}

\begin{table}[h]
	\centering
	\caption{The results of ablation study.}
	\begin{tabular}{lccccc}
		\hline
		Methods &~~AUC~~ &  ~~Dist~~  & ~~MDist~~ & ~~Ang~~ & ~~MAng~~ \\
		\hline
		Original image &0.839&0.212 & 0.146 & $32.6^{\circ}$ &$21.6^{\circ}$\\
		Original image + ROI head & 0.887& 0.182 & 0.118 & $22.9^{\circ}$ & $10.7^{\circ}$ \\
		W/O mid-layer supervision&0.875 & 0.178 & 0.101 & $24.4^{\circ}$ & $12.5^{\circ}$\\
		\hline
		Ours (one-scale)&\textbf{0.903} & \textbf{0.156} & \textbf{0.088} & $\mathbf{18.2^{\circ}}$ &$\mathbf{9.2^{\circ}}$\\
		\hline
		\hline
	\end{tabular}
	\label{Table:Ablation study}
\end{table}

The experiment results are listed in Table \ref{Table:Ablation study}. We can see that predicting heatmap merely based on the scene image is not easy, even the head and its position are already included in the image. With gaze direction as supervision (original image+ ROI head) to aid the heatmap pathway learning, the performance can be boosted. With a gaze direction pathway to predict gaze direction, our method greatly outperforms original image-based solution for gaze following, which further validates the importance of two-stage solution. Further the improvement of our method (one scale) over w/o mid-layer supervision validates the importance of gaze direction prediction, which is an advantage of our solution, \emph{i.e.,} our two-stage method benefits from gaze direction prediction.

\subsubsection{The information fusion.}
In the second stage, we combine the gaze direction field and image content information. However, how to choose the position (early, middle, late fusion) and way (multiplication or concatenation) of fusing? Here we compare our method with the following information fusion strategies:

\begin{itemize}
	\item Middle fusion (mul): Fuse the gaze direction field and the image content feature map ($7\times 7$) with multiplication in the middle layer. 
	\item Middle fusion (concat): Fuse the gaze direction field and the image content feature map ($7\times 7$) with concatenation in the middle layer. 
	\item Early fusion (mul): Fuse the gaze direction field and the image content feature map ($28 \times 28$) with multiplication in the early layer of encoder in heatmap pathway. 
	\item Late fusion (mul): Fuse the gaze direction field and the image content feature map ($28 \times 28$) with multiplication in the last layer of decoder in heatmap pathway. 
	\item Image fusion (mul): Directly multiply the original image with gaze direction field.
\end{itemize}

\begin{table}[h]
	\centering
	\caption{Different information fusion strategies.}
	\begin{tabular}{lccccc}
		\hline
		Methods &~~AUC~~ &  ~~Dist~~  & ~~MDist~~ & ~~Ang~~ & ~~MAng~~ \\
		\hline
		Middle fusion (mul) &0.882&0.183 & 0.118 & $21.7^{\circ}$ & $10.7^{\circ}$\\
		Middle fusion (concat) &0.884& 0.177 & 0.105 & $21.0^{\circ}$ & $10.5^{\circ}$\\
		Early fusion (mul) &0.898& 0.160 & 0.098 & $18.7^{\circ}$ & $9.6^{\circ}$\\
		Late fusion (mul) &0.888& 0.176 & 0.102 & $20.1^{\circ}$ &$10.1^{\circ}$\\
		Image fusion (mul) &0.895& 0.163 & 0.096 & $19.3^{\circ}$ & $9.7^{\circ}$ \\
		\hline
		Ours (concat) &\textbf{0.903} & \textbf{0.156} & \textbf{0.088} & $\mathbf{18.2^{\circ}}$ &$\mathbf{9.2^{\circ}}$\\
		\hline
		\hline
	\end{tabular}
	\label{Table:attention fusion}
\end{table}

Table \ref{Table:attention fusion} shows the results of different information fusion strategies. We can find that early fusion usually obtains higher performance than middle and late fusion, which implies early suppression of useless scene contents is important for gaze following. Furthermore, we find that usually concatenating the gaze direction field with image or feature achieves slightly better results than the multiplication. The possible reason is that the predicted gaze direction may not be very accurate, and the multiplication between image and the gaze direction field would lead to the change of intensities of pixels and cause information loss. While for concatenation, the information is still there and the heatmap pathway can tackle the heatmap prediction, even the gaze direction fields are not accurate.

\begin{table}[h]
	\centering
	\caption{The evaluation of different objectives.}
	\begin{tabular}{lccccc}
		\hline
		Methods &~~AUC~~ &  ~~Dist~~  & ~~MDist~~ & ~~Ang~~ & ~~MAng~~ \\
		\hline
		Point &0.892&0.173 & 0.103 & $21.9^{\circ}$ & $10.5^{\circ}$\\
		Multi-task point &0.900& 0.165 & 0.097 & $20.4^{\circ}$ &  $10.1^{\circ}$\\
		Shifted grid \cite{nips15_recasens} &0.899& 0.171 & 0.096 & $21.4^{\circ}$ & $10.3^{\circ}$ \\
		\hline
		Heatmap (our) &\textbf{0.903} & \textbf{0.156} & \textbf{0.088} & $\mathbf{18.2^{\circ}}$ & $\mathbf{9.2^{\circ}}$\\
		\hline
		\hline
	\end{tabular}
	\label{Table:output}
\end{table}

\subsubsection{Objective.}
Since the predicted gaze point may be multimodal, we introduce heatmap as the ground truth. Here, we also compare our method with networks based on other types of outputs, including:
\begin{itemize}
	\item Point: We employ two ResNet-50 to extract features for both original image and head image. Such a comparison is fair because the encoder part of FPN is also ResNet-50. In this baseline, we only predict gaze point.
	\item Multi-task regression: The network architecture is the same as point regression, but it predicts both gaze direction and gaze point simultaneously.
	\item Shifted grid: Based on our network architecture, shifted grid ($10\times 10$) \cite{nips15_recasens} is utilized to classify the gaze point into different grids.
\end{itemize}

\begin{figure*}[h]
	\centering
	\includegraphics[width=0.9\linewidth]{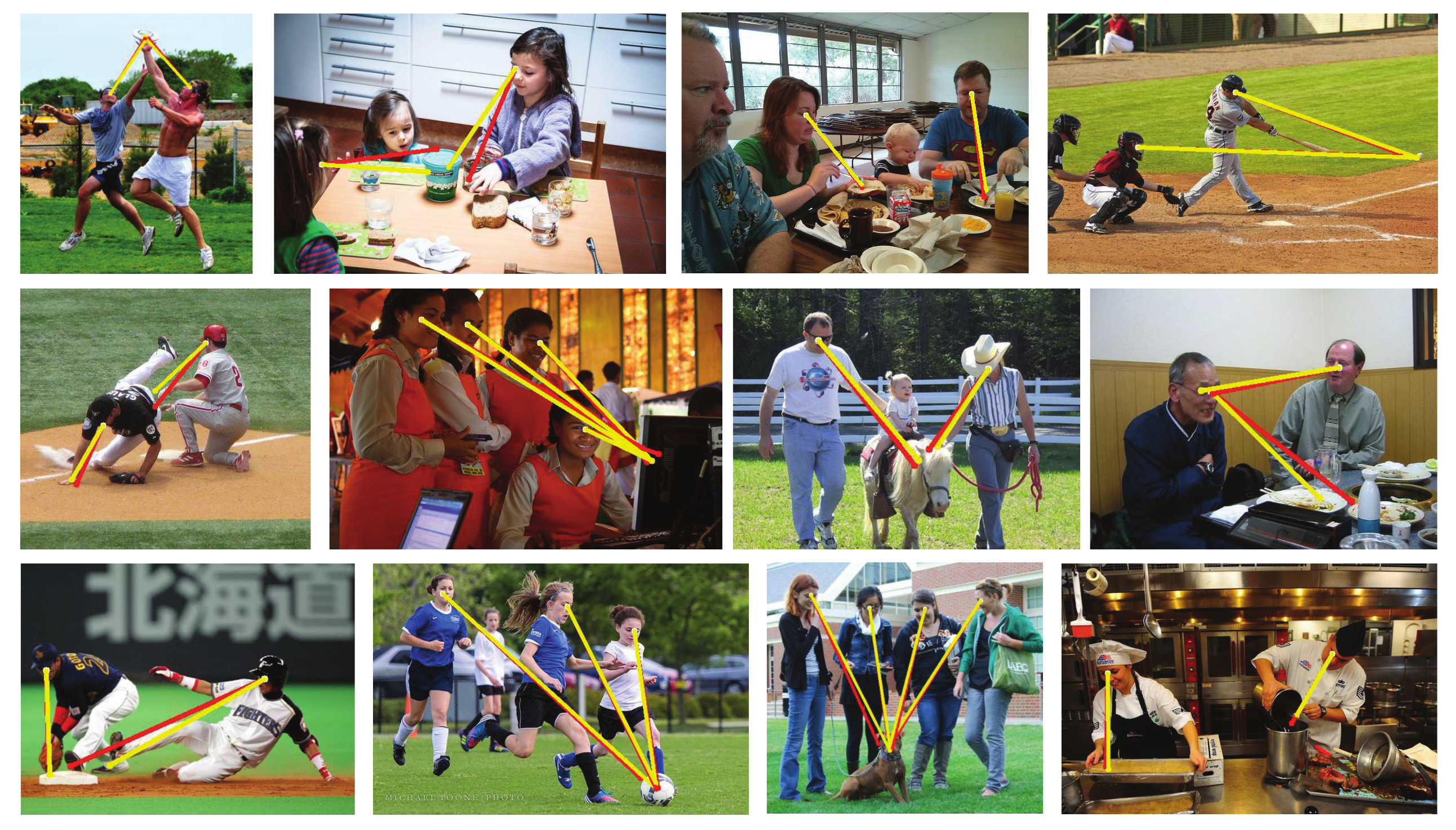}
	\caption{Some prediction results on the testing set, the red lines indicate the ground truth gaze and the yellow ones are the predicted gaze.}
	\label{fig:visualization_1} 
\end{figure*}

The comparison results of different objectives are listed in Table \ref{Table:output}. In our network architecture, heatmap regression achieves the best results than both point and shifted grid based objectives. As aforementioned, heatmap regression is more robust than directly point prediction because even a portion of heatmap values is incorrect, it is still possible to correctly predict the gaze point. Thus such a heatmap regression strategy is commonly used for human pose estimation \cite{fragkiadaki2015recurrent}. Our experiments also validate its effectiveness for gaze following.

\begin{figure}[h]
	\centering
	\subfigure[Some accurate preditions.]{
		\begin{minipage}[b]{0.47\textwidth}
			\includegraphics[width=1\linewidth]{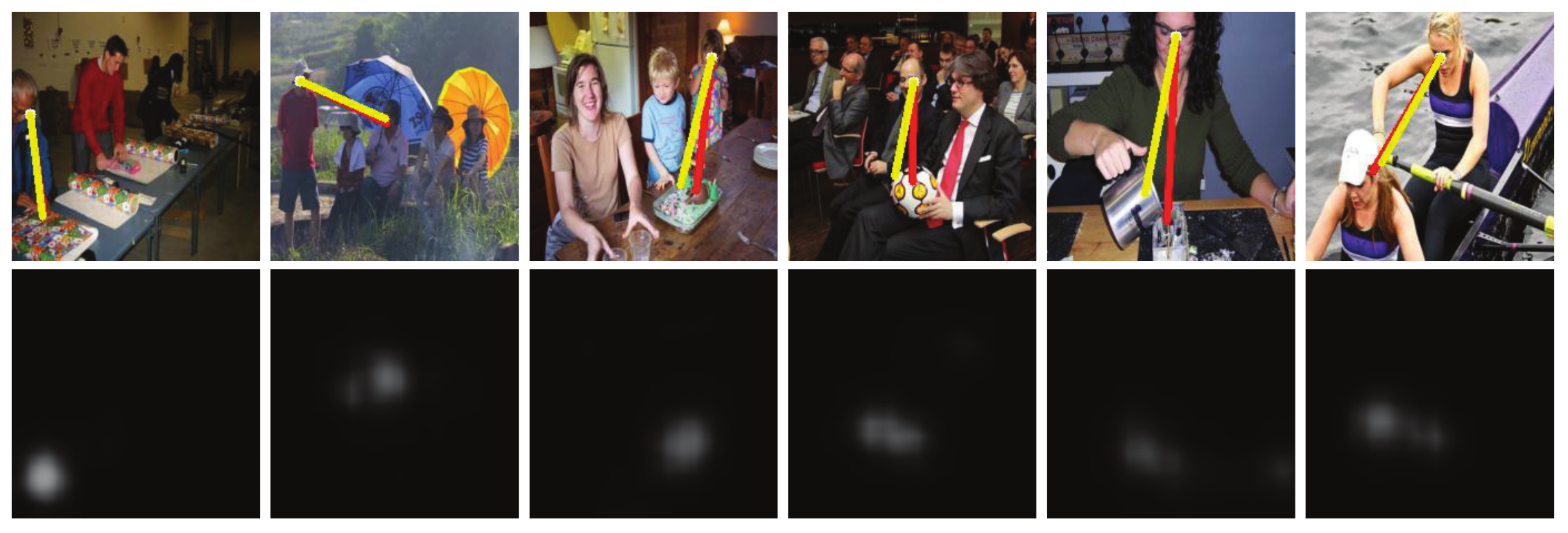}
		\end{minipage}
	}
	\subfigure[Some failures.]{
		\begin{minipage}[b]{0.47\textwidth}
			\includegraphics[width=1\linewidth]{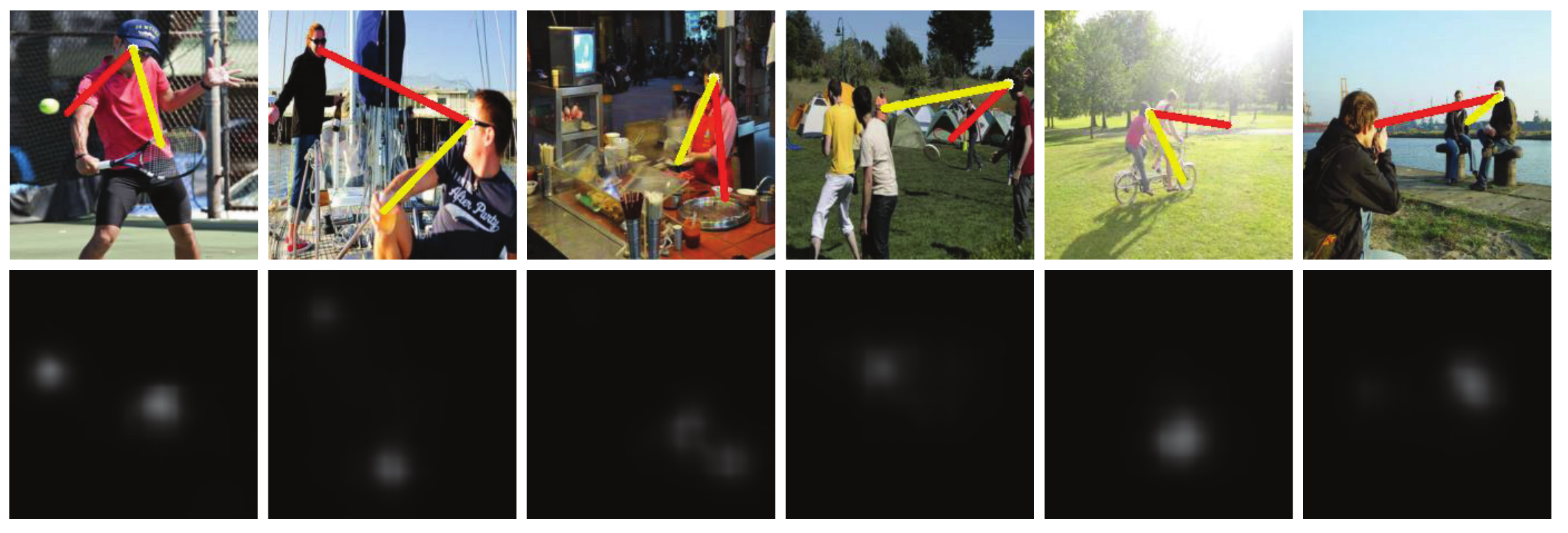}
		\end{minipage}
	}
	\caption{The first row: ground truth (red lines) and predicted gaze (yellow lines). The second row: predicted heatmaps. (Please zoom in for details.)}
	\label{fig:visualization_2}
\end{figure}

\subsection{Visualization of predicted results}
We show the predicted gaze points and their ground truth in Figure \ref{fig:visualization_1}, and predicted heatmaps in Figure \ref{fig:visualization_2} (a). We can see that for most of points, our method can predict gaze points accurately (As shown in Figure \ref{fig:accumulative}, when the distance error is 0.1, the portions of correctly predicted points is 50\% and 45\% on the GazeFollow dataset and DL Gaze, respectively.). There are two reasons contribute the good performance of our method: i) our two-stage solution agrees with the behavior of human, and gaze direction field would help suppress the regions falling out of gaze direction, which consequently improves the heatmap regression; ii) the supervision on gaze direction helps train a more robust network for gaze following.
We also show some failures in Figure \ref{fig:visualization_2} (b). The first three columns of examples show that our predictions can be multimodal. Although the position of heatmap maximum is not right, some others peaks can also predict the gaze point. Regarding the last three columns of failures, we can see that the predicted heatmap is inaccurate. This probably caused by the small head or head occlusion, which makes gaze direction and gaze point prediction extremely difficult, even for us human.

\section{Conclusion}
\label{sec:conclusion}
In this paper, we proposed a two-stage solution for gaze tracking. In stage I, we feed the head image and its position for gaze direction prediction. Then we use gaze direction field to characterize the distribution of gaze points without considering the scene contents. In stage II, the gaze direction fields are concatenated with original image, and fed into a heatmap pathway for heatmap regression. The advantages of our solution are two-fold: i) our solution mimics the behavior of human in gaze following, therefore it is more psychological plausible; ii) besides leverage heatmap to supervise the training of our network, we can also leverage gaze direction to facilitate the training of gaze direction pathway, therefore our network can be more robustly trained. We further build a new DL Gaze dataset to validate the performance of different gaze following methods in real scenarios. Comprehensive experiments show that our method significantly outperforms existing methods, which validates the effectiveness of our solution. 
\\\\
\noindent\textbf{\large Acknowledgement.} 
This project is supported by NSFC (No. 61502304).

\clearpage
\bibliographystyle{splncs}
\bibliography{egbib}

\title{Supplementary Material} % Replace your paper's title here
\titlerunning{Believe It or Not, We Know What You Are Looking at!} % Replace an abstracted version of your paper's title here

%===========================================================

\author{Dongze Lian\thanks{The authors contribute equally.}\orcidID{0000-0002-4947-0316} \and
	Zehao Yu$^*$\orcidID{0000-0002-6559-9830} \and
	Shenghua Gao\thanks{Corresponding author.}\orcidID{0000-0003-1626-2040}}
%\author{}

\authorrunning{Lian et al.}

\institute{School of Information Science and Technology, ShanghaiTech University\\
	\email{\{liandz, yuzh, gaoshh\}@shanghaitech.edu.cn}}

\maketitle
This supplementary material provides some supplementary notes and results. First, we give the detailed information of evaluation metric in Section 4.1 (\textbf{AUC}, \textbf{Dist}, \textbf{MDist}, \textbf{Ang}). Secondly, we analyze the running time of these methods. Finally, we explain the choice of parameter $\gamma$ in the Eq. 3. 

\section{Evaluation metric}

\textbf{Area Under Curve (AUC)}: The area under ROC curve, which is generated according to \cite{judd2009learning}.

\noindent\textbf{$L_2$ distance (Dist)}: The Euclidean distance between predicted gaze point and the average of ground truth annotations. The original image size is normalized to 1 $\times$ 1.

\noindent\textbf{Minimum $L_2$ distance (MDist)}: The minimum Euclidean distance between predicted gaze point and all ground truth annotations. The original image size is normalized to 1 $\times$ 1.

\noindent\textbf{Angular error (Ang)}: The angular error between predicted gaze direction and ground truth direction corresponding to average gaze point.

\section{Running time}
We compare the performance and the running time of different baselines in the testing phase. All the methods are tested with an NVIDIA Titan X GPU. The metric used here is the $L2$ distance between predicted gaze point and the average of ground truth annotation. The running time refers to the running time for one testing sample, which is an average running time of 1000 test samples (The unit is millisecond). We repeat the experiments ten times.

\section{The choice of $\gamma$}
If the predicted gaze direction is accurate, it is desirable that the probability distribution is sharp along the $\theta$, otherwise, it is desirable that the probability changes smoothly. Thus, we choose three different $\gamma$ to represent the decay rate in the Eq. 3 and our experiments verify the effectiveness of multi-scale gaze direction field. About the choice of $\gamma$, considering the decay rate, we empirically set $\gamma$ to 1, 2, 5, respectively. That is because when the value of the cosine function is $0.5$, the angle is about $60^{\circ}$, $45^{\circ}$ and $30^{\circ}$. The gaze direction fields of different $\gamma$ are shown in Fig. 2 in the paper.

\begin{table}[h]
	\centering
	\caption{The performance and running time of different methods on GazeFollow.}
	\begin{tabular}{lcccc}
		\hline
		Methods  &  ~~Dist~~  & ~~time (ms)~~ \\
		\hline
		SalGAN \cite{pan2017salgan} & 0.238 & 8.8\\
		Recasens \emph{et al.} \cite{nips15_recasens} & 0.190 & 10.4 \\
		Recasens* \emph{et al.} \cite{nips15_recasens} & 0.175 & 15.8 \\
		\hline
		Our method (one-scale) & 0.156 & 16.1\\
		Our method (multi-scale) & 0.145  & 16.3\\
		\hline
		\hline
	\end{tabular}
	\label{Table:time}
\end{table}

\end{document}